\documentclass{article} 
\usepackage{iclr2024_conference,times}

\usepackage{amsmath,amsfonts,bm}

\def\eqref#1{equation~\ref{#1}}

\def\1{\bm{1}}

\DeclareMathAlphabet{\mathsfit}{\encodingdefault}{\sfdefault}{m}{sl}
\SetMathAlphabet{\mathsfit}{bold}{\encodingdefault}{\sfdefault}{bx}{n}

\usepackage{hyperref}
\usepackage{url}
\usepackage{multirow}
\usepackage{graphicx}
\usepackage{subfigure}
\usepackage{wrapfig}

\title{Interpretable Neural Temporal Point Processes For Modelling Electronic Health Records}

\author{Bingqing Liu \\
University of Chinese Academy of Science; Academy of Mathematics and System Science, CAS\\
\texttt{liubingqing20@mails.ucas.ac.cn} \\
}

\iclrfinalcopy 
\begin{document}

\maketitle

\begin{abstract}
Electronic Health Records (EHR) can be represented as temporal sequences that record the events (medical visits) from patients. Neural temporal point process (NTPP) has achieved great success in modeling event sequences that occur in continuous time space. However, due to the black-box nature of neural networks, existing NTPP models fall short in explaining the dependencies between different event types. In this paper, inspired by word2vec and Hawkes process, we propose an interpretable framework inf2vec for event sequence modelling, where the event influences are directly parameterized and can be learned end-to-end. In the experiment, we demonstrate the superiority of our model on event prediction as well as type-type influences learning.
\end{abstract}

\section{Introduction}
\label{sec:intro}
Event sequence is a ubiquitous data structure in real world, such as user behavior sequences, error logs, 
purchase transaction records and electronic health records \citep{Mannila1997DiscoveryOF,Liu1998IntegratingCA,Zhou2013LearningSI,Choi2016RETAINAI,Liu2023LinkAwareLP}. An event can be generally represented as a tuple, including the event type and occurrence time, e.g., (well-child visit, 2024/02/01). Inside the event sequence, various types of events often exhibit complex temporal patterns, making the type-type influences discovering even challenging. Temporal point process has been a popular and principled tool for event sequence modeling \cite{shchur2021neural}. Due to the high capacity of deep networks, neural temporal point process models have been intensively devised and have demonstrated superior performance for tasks such as event prediction \citep{du2016recurrent,omi2019fully,waghmare2022modeling,soen2021unipoint,Zhou2023AutomaticIF,mei2017neural,chen2018neural}. However, their black-box nature makes most of them lack transparency and prevents them from explaining their decisions \citep{danilevsky2020survey,minh2022explainable,linardatos2020explainable}. 

Some attempts are made to make the event sequence model more explainable. AutoNPP \citep{Zhou2023AutomaticIF} adopts the additive form of the intensity function to capture the historical impacts, like Hawkes process. NRI-TPP \citep{Zhang2021NeuralRI} leverages the variational inference to recover the underlying event dependencies by message passing graph and recurrent neural network (RNN). CAUSE \citep{Zhang2020CAUSELG} learns the Granger causality between event types by attribution methods, namely integrated gradients. Attention mechanism is also widely used \citep{zuo2020transformer,zhang2020self,Choi2016RETAINAI, dash2022learning,Gu2021AttentiveNP}. However, these models either trade accuracy and efficiency for interpretability, or are only designed for discovering the token-wise influcences, or are model-specific, i.e., can not be applied to other models. Moreover, for the mostly widely used attention mechanism, there exists layer and head inconsistency about its interpretability, i.e., different attention layers and heads have different attention scores, which undermines its practical utility. To learn type-type influences, we draw inspiration from word2vec \citep{mikolov2013distributed}, which learns semantic vector representations that can help group similar words. Though this vanilla interpretability can not indicate the type-type influences, the mutual influences could be reflected if we take a further step: create a vector space for each event type. As a result, distributed representations of event types in the vector space of event type $k$ can help group event types that have similar influences on $k$.


\section{Method}
\label{sec:method}

\begin{figure*}
\centering 
\includegraphics[width=0.85\textwidth]{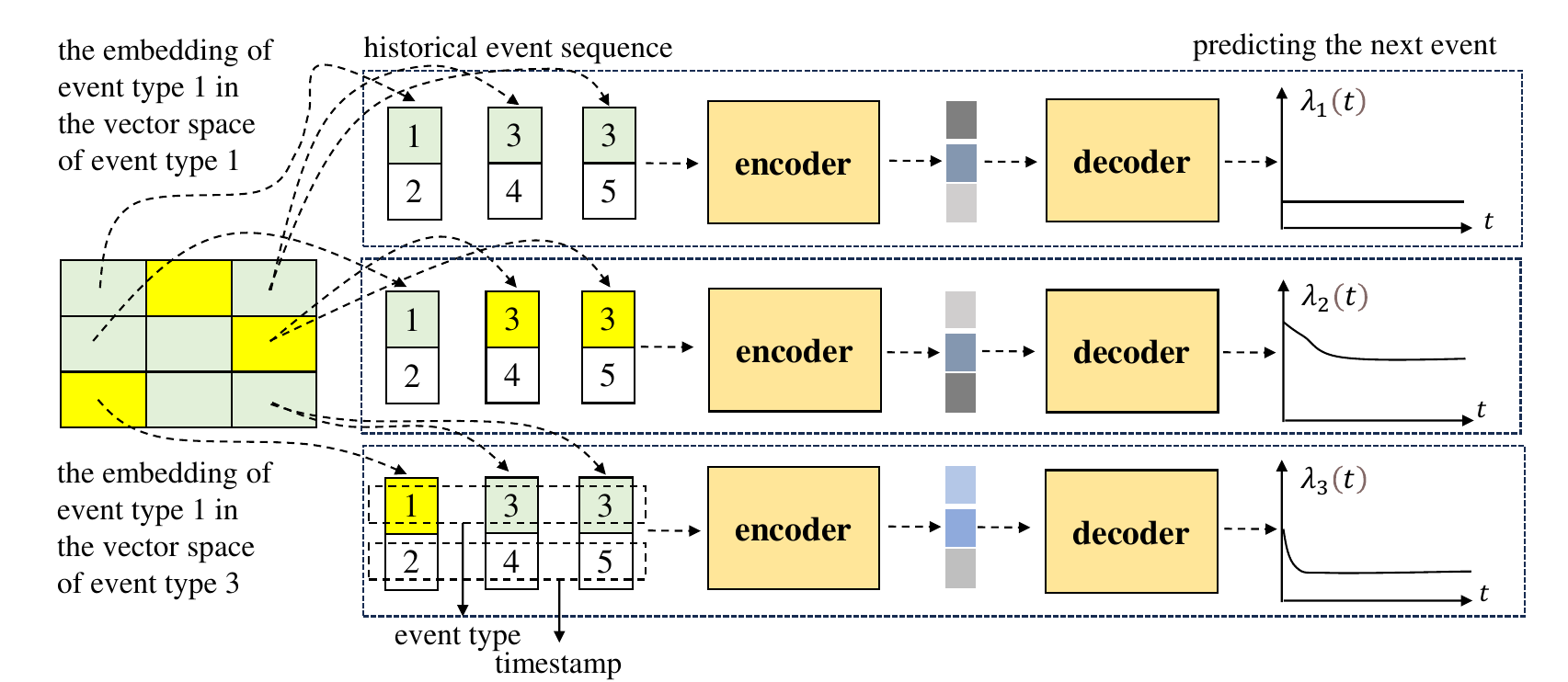}
\caption{An overview of our proposed type-type influences learning framework. The framework creates separate embedding, encoding and decoding space for each event type. In the illustrated example, brighter color in the embedding layer means stronger influence and we see event of type 2 is more likely to occur.}
\label{fig:model}
\end{figure*}

In this section, we detail the proposed type-type influences learning framework, namely inf2vec. The event sequence model has three modules: the embedding layer, the sequence encoder, and the event decoder. Besides word2vec, our model also draws inspiration from Hawkes process \citep{hawkes1971point,hawkes1971spectra}. Both of us model the event dynamics as an influence-driven process in the view of event types. To capture the impacts of historical events $\{(k_i,t_i)|t_i<t\}$,  Hawkes process specifies the conditional intensity function as follows,
\begin{equation}
\label{eq:mhp}
\lambda_k(t)=\mu_k + \sum \limits_{t_i<t} \alpha_{k,k_i} \exp(-\beta_{k,k_i}(t-t_i))
\end{equation}
where $\mu_k\ge0$ is the base intensity, $\alpha_{k,k_i}\ge0$ is the coefficient indicating how significantly event $k_i$ will influence the occurrence of event $k$, and $\beta_{k,k_i}\ge0$ shows how the influence decays over time. The schematic representation of our model is illustrated in Fig.\ref{fig:model}.

\subsection{Embedding Layer}
\label{subsec:emb}
\textbf{\textit{Global vs local embedding.}} In conventional neural temporal point process models, the embedding layer assigns each event type a vector representation and the learned type representations can naturally group similar event types. However, it's hard for this kind of embedding to explain the dependencies between different event types. For example, which event type influences the given event type $k$ most? One may turn to techniques like dot product to find the most influencing event type but again, it can only find the most similar rather than influencing event type. In fact, the conventional embedding can be considered as the global embedding, i.e., the vector representations of different types are in the same vector space. Our idea is to create $K$ vector spaces and in each, an event type will have a vector representation, which we call local embedding. This kind of embedding can naturally reflect the relationships between different event types. An event type can have different vector representations in different vector spaces, indicating that it can have different influences on different event types. Moreover, given the vector space of event type $k$, close vector representations in this space indicate that the corresponding event types have similar impacts on event type $k$. 

The local embedding is inspired by Hawkes process (Eq.\ref{eq:mhp}), where the impact of historical event $(k_i,t_i)$ on the occurrence of event $k$ is explicitly characterized by two learnable parameters $\alpha_{k,k_i}$ and $\beta_{k,k_i}$. From the perspective of embedding, the two parameters $[\alpha_{k,k_i},\beta_{k,k_i}]$ can be considered as the vector representation of event type $k_i$ in the vector space of event type $k$, with the embedding dimension being 2. More generally in this work, we set the embedding dimension as a hyperparameter $d$ and use notation $\boldsymbol{z_m^k}(k_i)\in \mathbb{R}^d$ to denote the embedding of event type $k_i$ in the vector space of event type $k$. And we use temporal embedding function $\boldsymbol{z_t^k}(t_i)$ to embed the timestamp $t_i$. Then in the context of event type $k$, we obtain the event embedding by concatenating the type and time embedding,
\begin{equation}
\label{eq:event}
	\begin{aligned}
\boldsymbol{e^k}(i)&=\boldsymbol{z_m^k}(k_i)||\boldsymbol{z_t^k}(t_i)
	\end{aligned}
\end{equation}

\subsection{Sequence Encoder}
\textbf{\textit{Global vs local encoding.}}  To capture the impacts of historical events $\{(k_j,t_j)\}_{j=1}^i$, existing sequence encoders encode them into one singe vector, which can be summarized as follows:
\begin{equation}
\label{eq:prev_enc}
	\begin{aligned}
\boldsymbol{h_i}&=Seq2Vec(\boldsymbol{e}(1)\cdots,\boldsymbol{e}(j),\cdots,\boldsymbol{e}(i))
	\end{aligned}
\end{equation}
where  $\boldsymbol{e}(j)$ is the embedding of event $(k_j,t_j)$ and the backbone network $Seq2Vec$ is usually realized by RNN \citep{chung2014empirical} or Transformer \citep{vaswani2017attention}. The conventional encoder is devised specifically for the global event embedding, which we call global encoding. To handle the local event embedding, we here propose type-wise encoder:
\begin{equation}
\label{eq:tw_enc}
	\begin{aligned}
\boldsymbol{h_i^k}&=Seq2Vec^{k}(\boldsymbol{e^k}(1)\cdots,\boldsymbol{e^k}(j),\cdots,\boldsymbol{e^k}(i))
	\end{aligned}
\end{equation}
which performs separate history encoding for different event types. The underlying rationale for the type-wise encoder is that each event type can concentrate on the historical events of its own interests, which we call local encoding. For example, if only the event $(k_j,t_j)$ has impact on the occurrence of event $k$, then the history encoding $\boldsymbol{h_i^k}$ can only encode the information from $\boldsymbol{e^k}(j)$ and ignore the other uninterested events. As a comparison, the conventional encoder tries to summarize all historical information into one single vector $\boldsymbol{h_i}$.  From this point of view, we can regard the local encoding $\{\boldsymbol{h_i^k}\}_{k=1}^K$ as the information decoupling (from the perspective of event types) of the global encoding $\boldsymbol{h_i}$, with each local encoding $\boldsymbol{h_i^k}$ only containing the historical information that the event type $k$ is interested in. 

\subsection{Event Decoder}
\textbf{\textit{Global vs local decoding.}}  With history representation obtained from the sequence encoder, we are about to decode the next event (the $(i+1)\mbox{-}th$ event). In existing neural temporal point process models, the next event distribution is mostly characterized by the conditional intensity function, which we summarize as follows,
\begin{equation}
\label{eq:dec_int}
	\begin{aligned}
\lambda_k(t)&=\sigma(NN_k(\boldsymbol{h_i},t))
	\end{aligned}
\end{equation}
where $\sigma$ is an activation function to ensure the positive constraint of the intensity function and $NN_k$ is a neural network, e.g., multi-layer perceptron. We see that in the conventional decoder, the global history encoding $\boldsymbol{h_i}$ is shared across different intensity decoders $\{NN_k\}_{k=1}^K$, which we call global decoding. Each intensity decoder takes out the information of interests from the global history encoding $\boldsymbol{h_i}$ and generate the corresponding intensity $\lambda_k(t)$. But in our framework, extracting information of interests is unnecessary as the global history encoding has already been decoupled into local history encoding in the view of event types in the encoding stage. Therefore, we can simply replace the global history encoding as the local history encoding, i.e.,
\begin{equation}
\label{eq:dec_int_deer}
	\begin{aligned}
\lambda_k(t)&=\sigma(NN_k(\boldsymbol{h_i^k},t))
	\end{aligned}
\end{equation}
we call it local decoding. In temporal point process, the next event distribution can be also described by cumulative hazard function \citep{omi2019fully}, probability density function \citep{shchur2021neural}, etc. They have quite different functional formulations comparing to Eq.\ref{eq:dec_int}. But without loss of generality, we can accordingly adapt them to our framework. 

\section{Experiment}
In this part, we design experiments to answer the following
questions: \textbf{Q1}, What's the performance of inf2vec on standard prediction tasks? \textbf{Q2}, How's the quality of the learned type-type influences? 
\subsection{Datasets} Three publicly available EHR datasets are used, namely SynEHR1,  SynEHR2, and MIMIC \citep{Enguehard2020NeuralTP,waghmare2022modeling}. Each dataset has a number of event sequences and each sequence records the medical events from a patient. They have 6, 178, and 75 event types, respectively. To evaluate the quality of the learned type-type influences, we additionally use three synthetic datasets, Haw5, Haw9 and HawC9, which are all simulated by Hawkes process (Eq.\ref{eq:mhp}) but have different parametric settings. Technically, we use the open-source python library tick  \footnote{https://x-datainitiative.github.io/tick/modules/hawkes.html} to simulate the Hawkes process. These three Hawkes datasets have 5, 9 and 9 event types, respectively. Each dataset is split into training, validation, and testing data according the number of event sequences, with each part accouting for 60\%, 20\% and 20\%, respectively.

\subsection{Compared Models and Implemented inf2vec}
The compared models are state-of-the-art baselines. RNN based models include NHP \citep{mei2017neural}, FullyNN \citep{omi2019fully} and JTPP \citep{waghmare2022modeling}. Transformer based models include SAHP \citep{zhang2020self} and THP \citep{zuo2020transformer}. In the decoding stage, NHP, SAHP and THP model the intensity function, FullyNN models the cumulative hazard function and JTPP models the probability density function. We adapt JTPP to our framework, i.e., to implement inf2vec, we use RNN as our encoder, characterize the time distribution using probability density function, and train the model by maximum likelihood estimation. To reduce parameters, one single RNN ($Seq2vec$) is shared across different event types (Eq.\ref{eq:tw_enc}). 
\subsection{Prediction Results}
The predictive ability of one NTPP model can be evaluated by predicting the next event, including the event type and occurrence time, given the historical events.  As datasets used exhibit class (type) imbalance, we use weight F1 score to report the accuracy of event type prediction. For time prediction, we use mean absolute error as the evaluation metric. The results for the five datasets are summarized in Table \ref{table:nlltype} (Haw5 is not compared due to space limit). We can see that inf2vec outperforms the baselines in all datasets (except for JTPP). We use JTPP's encoder and decoder, and adapt them to our framework. We see the implemented inf2vec performs better than JTPP in most datasets. Compared with these baselines, our model conducts information decoupling in the view of event types, gaining more effectiveness in events embedding, encoding and decoding.

\begin{table*}
\caption{Comparison of weighted F1 score and mean absolute error on event prediction. The results are averaged by 10 runs and we use bold numbers to indicate the best performance.}
\centering
\begin{tabular}{c|cc|cc|cc|cc|cc}
 \hline
 \multirow{2}{*}{Model} & \multicolumn{2}{c|}{Haw9} & \multicolumn{2}{c|}{HawC9}& \multicolumn{2}{c|}{SynEHR1}& \multicolumn{2}{c|}{SynEHR2}& \multicolumn{2}{c}{MIMIC}\\
\cline{2-11} 
&F1&MAE& F1&MAE& F1&MAE& F1&MAE& F1&MAE\\
 \hline
NHP&22.8&0.46&36.5&0.41&56.3&1.93&53.3&3.11&62.3&0.41\\
FullyNN&22.6&0.40&36.2&0.38&59.4&1.35&56.0&2.42&63.9&0.30\\
SAHP&22.5&0.44&36.3&0.39&58.9&1.85&55.7&2.81&63.3&0.33\\
THP&23.7&0.43&37.4&0.39&58.7&1.77&55.3&2.73&63.5&0.35\\
JTPP&23.1&0.39&38.5&\textbf{0.34}&59.6&\textbf{1.05}&56.1&2.11&64.9&0.27\\
\hline
\textbf{Inf2vec} &\textbf{24.5}&\textbf{0.37}&\textbf{39.1}&0.36&\textbf{60.2}&1.11&\textbf{56.4}&\textbf{2.01}&\textbf{65.3}&\textbf{0.24}\\
\hline
\end{tabular}
\label{table:nlltype}
\end{table*}
\begin{figure}
\centering 
\includegraphics[width=1\textwidth]{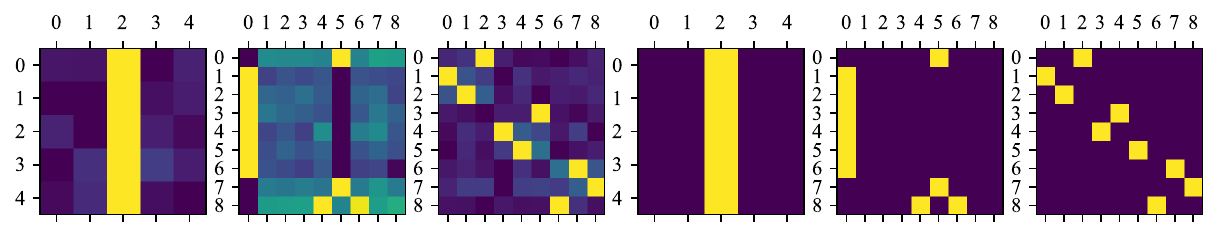}
\caption{The illustration of our learned local embeddings (after dimension reduction) over dataset Haw5, Haw9 and HawC9 (the first three plots), coordinates $(x,k)$ denotes the embedding of event type $x$ in the context of event type $k$. The last three plots are the ground truth influences and coordinates $(x,k)$ denotes the influence of $x$ on $k$, brighter color means stronger influence.}
\label{fig:dependency_haw9c9}
\end{figure}

\subsection{Type-Type Influences Learning} \begin{wrapfigure}{r}{.5\textwidth}
  \vskip -\baselineskip
  \includegraphics[width=\linewidth]{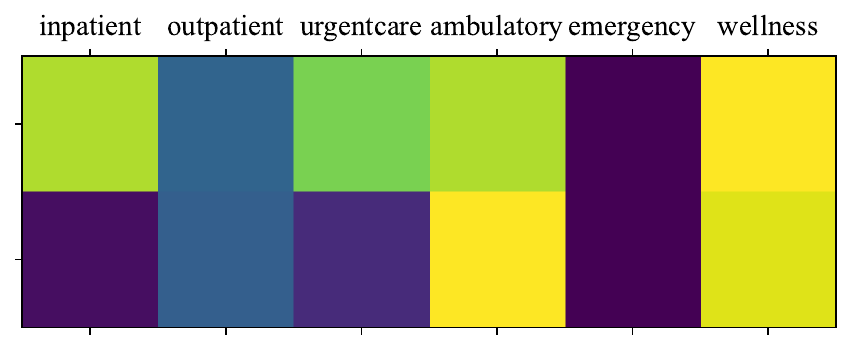}
  \caption{The illustration of our learned local embeddings over dataset SynEHR1. The first and second row show how other events influence "wellness" and "ambulatory", respectively. Brighter color indicates stronger influence.}
  \label{fig:synehr1}
\end{wrapfigure} In Hawkes process (Eq.\ref{eq:mhp}), the coefficients $\alpha_{k,*}$ and $\beta_{k,*}$ indicate how other event types influence event type $k$, against which we can compare our learned type-type influences. In our framework inf2vec, the local embedding $\boldsymbol{z_m^k}(k_i)$ can naturally reflect the relationship between event type $k_i$ and $k$. And in the vector space of event type $k$, close vector representations indicate that the corresponding event types have similar impacts on event type $k$. To evaluate the quality of type-type influences learning of our framework, for our learned local embeddings $\boldsymbol{z_m^k}(*)\in \mathbb{R}^d$ and the ground truth influences $[\alpha_{k,*},\beta_{k,*}]\in \mathbb{R}^2$, we reduce the vector dimension to 1 by techniques like principle component analysis (PCA) \cite{abdi2010principal} and visualize them. Fig.\ref{fig:dependency_haw9c9} illustrates the results on the three synthetic Hawkes datasets, showing that our learned local embeddings exhibit a high consistency with the ground truth influences, which validates the effectiveness of our proposed framework. Over EHR dataset SynEHR1, we show the learned event influences for event type "wellness" and "ambulatory" in Fig.\ref{fig:synehr1}. The results show that our learned event influences are mostly consistent with human experience.

\section{Conclusion}
In this paper, we present a type-type influences learning framework Inf2vec for neural temporal point process, where the influences are directly parameterized and are learned end-to-end. Compared with conventional NTPP approaches, our framework conducts information decoupling from the perspective of event types, leading to more efficient embedding, encoding and decoding. Our framework is quite general for not posing any restriction on model's encoder and decoder architecture. Experimental results on both synthetic and real-world EHR datasets  demonstrate the superior performance of our model in terms of event prediction and influence learning task. 

\bibliography{iclr2024_conference}
\bibliographystyle{iclr2024_conference}


\end{document}